\RequirePackage{fix-cm}
\documentclass[smallextended]{svjour3}       % onecolumn (second format)
\smartqed  % flush right qed marks, e.g. at end of proof
\usepackage{graphicx}

\usepackage[misc]{ifsym} % Letter for corresponding author

\usepackage{algorithmicx,algorithm}

\usepackage{multirow}
\newcommand{\tabincell}[2]{\begin{tabular}{@{}#1@{}}#2\end{tabular}}

\usepackage{lineno,hyperref}
\usepackage{setspace}
\usepackage{amssymb}
\usepackage{verbatim}

\usepackage[numbers,sort&compress]{natbib}

\begin{document}

\title{Dual Discriminator Adversarial Distillation for Data-free Model Compression%\thanks{Grants or other notes
%about the article that should go on the front page should be
%placed here. General acknowledgments should be placed at the end of the article.}
}
\subtitle{}

%\titlerunning{Short form of title}        % if too long for running head

\author{{Haoran Zhao}    \textsuperscript{1}   \and
        {Xin Sun}   \textsuperscript{1,2~\Letter}   \and
        {Junyu Dong}    \textsuperscript{1~\Letter}  \and
        {Milos Manic}   \textsuperscript{3}   \and
        {Huiyu Zhou}    \textsuperscript{4}   \and
        {Hui Yu}    \textsuperscript{5}   
}

%\authorrunning{Short form of author list} % if too long for running head

\institute{%
    \begin{itemize}
      \item[\textsuperscript{\Letter}] {Xin Sun} \\
            \email{sunxin@ouc.edu.cn}
      \item[\textsuperscript{\Letter}] {Junyu Dong} \\
            \email{dongjunyu@ouc.edu.cn}
      \at
      \item[\textsuperscript{1}] College of Information Science and Engineering, Ocean University of China, Qingdao, PR China
      \item[\textsuperscript{2}] Department of Aerospace and Geodesy, Technical University of Munich, Germany
      \item[\textsuperscript{3}] College of Engineering, Virginia Commonwealth University, US
      \item[\textsuperscript{4}] School of Informatics, University of Leicester, UK
      \item[\textsuperscript{5}] School of Creative Technologies, University of Portsmouth, Portsmouth, UK
    \end{itemize}
}

\date{Received: date / Accepted: date}
% The correct dates will be entered by the editor
\maketitle

\begin{abstract}
Knowledge distillation has been widely used to produce portable and efficient neural networks which can be well applied on edge devices for computer vision tasks. However, almost all top-performing knowledge distillation methods need to access the original training data, which usually has a huge size and is often unavailable. To tackle this problem, we propose a novel data-free approach in this paper, named Dual Discriminator Adversarial Distillation (DDAD) to distill a neural network without the need of any training data or meta-data. To be specific, we use a generator to create samples through dual discriminator adversarial distillation, which mimics the original training data. The generator not only uses the pre-trained teacher's intrinsic statistics in existing batch normalization layers but also obtains the maximum discrepancy from the student model. Then the generated samples are used to train the compact student network under the supervision of the teacher. The proposed method obtains an efficient student network which closely approximates its teacher network, without using the original training data. Extensive experiments are conducted to demonstrate the effectiveness of the proposed approach on CIFAR, Caltech101 and ImageNet datasets for classification tasks. Moreover, we extend our method to semantic segmentation tasks on several public datasets such as CamVid, NYUv2, Cityscapes and VOC 2012. To the best of our knowledge, this is the first work on generative model based data-free knowledge distillation on large-scale datasets such as ImageNet, Cityscapes and VOC 2012. Experiments show that our method outperforms all baselines for data-free knowledge distillation.
\keywords{Deep neural networks \and image classification \and model compression \and knowledge distillation \and data-free}

\end{abstract}

\section{Introduction}
\label{intro}
Deep convolutional neural networks (CNNs) have attained state-of-the-art performance in various computer vision tasks, e.g., image classification
\cite{He_2016_CVPR,9293172,9382121,9027090}, object detection \cite{DBLP:conf/nips/RenHGS15,DBLP:journals/tmm/LouWNHMWY20,9234727,9245532} and semantic segmentation \cite{DBLP:conf/cvpr/LongSD15,MING202114,Chen_Sun_Hua_Dong_Xv_2020,Zhang,9345439}. Whereas, along with high-performance, CNNs 
with tens of millions of parameters, hundreds of layers often require heavy computation and storage. Thus, one prominent obstacle is that it is impossible to deploy such complicated CNNs into resource-constrained devices, such as mobile-phone and autonomous cars.

To this end, a large body of works have been published \cite{9292449,8816693,9351789,8895819,9044324,9126831}, which intend to compress and speed-up the cumbersome CNNs into more lightweight ones. The popular compression techniques are divided into five categories: quantization \cite{DBLP:conf/icml/GuptaAGN15,9082051}, low rank factorization, pruning \cite{DBLP:conf/nips/HanPTD15,9097925} and sharing weights, compact convolutional filter, and knowledge distillation \cite{Hinton2015Distilling}. For example, Guo et al. \cite{9097925} propose a novel pruning framework which can compress the cumbersome model by pruning the channels and automatically decide the network structure after channel pruning. Kang et al. \cite{8693518} introduce a novel pruning framework that reflects accelerator architectures. Liu et al. \cite{9357413} propose to compress a compact student by distilling the cross-modal representations of the pre-trained multi-modal teacher. Among these methods, knowledge distillation \cite{Hinton2015Distilling} is one of the most popular paradigms for learning a portable student model from the pre-trained complicated teacher by directly imitating its outputs.

Knowledge distillation (KD) is proposed by Hinton et al.  \cite{Hinton2015Distilling} for supervising the training of a compact yet efficient student model by capturing and transferring the knowledge of a large teacher model to a compact one. Note that, the student model pays more attention to the probability correlation between categories rather than the one-hot labels by softening the outputs of the teacher in softmax. Further researches propose various kinds of distillation loss functions in the field of CNNs optimization. For example, FitNets \cite{Romero2015FitNets} introduces the hints loss to guide the training of intermediate layers of the student network. AT ~\cite{Zagoruyko2016Paying} transfers the attention features of intermediate layers from the teacher network to student network. FSP \cite{Yim2017A} designs the flow distillation loss to force the student model to mimic flow matrices of the teacher model among the feature maps between two layers. RKD \cite{park2019relational} transfers mutual relations of data examples by the distance-wise and angle-wise distillation losses. SP \cite{tung2019similarity} preserves the pairwise similarities in student's representation space instead to mimic the representation space of the teacher.

The aforementioned knowledge distillation approaches have achieved high efficiency and accuracy on public datasets when the original training dataset can be obtained. However, we cannot access the original training dataset in many scenarios such as security and privacy. Thus, we are not able to directly use these existing knowledge distillation approaches for distilling a compact student model.

Nevertheless, a few works \cite{chen2019data,micaelli2019zero,DBLP:conf/icml/NayakMSRC19,DBLP:journals/corr/abs-1710-07535,DBLP:conf/nips/YooCKK19,DBLP:conf/cvpr/YinMALMHJK20,DBLP:journals/corr/abs-1912-11006} have been proposed to obtain the portable student model by knowledge distillation without the need of any original training data. For example, Chen et al. \cite{chen2019data} and Micaelli et al. \cite{micaelli2019zero} conduct pilot research on data-free knowledge distillation. In Chen's work \cite{chen2019data}, the one-hot samples are generated to replace original samples, which can highly activate the neurons of the teacher network. Micaelli et al. \cite{micaelli2019zero} propose to search images that poorly match the teacher network. Besides, Lopes et al. \cite{DBLP:journals/corr/abs-1710-07535} utilize the layer spectral activations or the layer activations from the teacher model to reconstruct the training samples. Nayak et al. \cite{DBLP:conf/icml/NayakMSRC19} propose to model the softmax space using the parameters of the teacher model to transfer data. However, these  methods only take account of information from the student but ignores the valuable information of the pre-trained teacher model. Besides, there has been another perspective \cite{DBLP:conf/cvpr/YinMALMHJK20} which starts with an arbitrary input and is iteratively updated via back-propagation. Nayak et al. \cite{DBLP:conf/icml/NayakMSRC19} synthesize the training samples named Data Impressions via modelling the data distribution in the softmax space. Another two related works \cite{DBLP:conf/cvpr/YinMALMHJK20,DBLP:conf/cvpr/HaroushHHS20} extend the Inceptionism method \cite{mordvintsev2015inceptionism} with some regularizers. Although the generated images from these approaches  are realistic, they leave the student information constraint-free. 

In summary, works mentioned above have obtained impressive results on classification tasks but still suffer from several limitations. One kind of data-free knowledge distillation methods, which exploit the generative adversarial networks, only aim to train the generator for synthesizing images that could fool the student. Another kind of works which update the arbitrary input via back-propagation, only take the pre-trained teacher model into account but ignore the information from the student. The generated samples obtained by existing works cannot well balance the realistic of images and the generalization of student model.
Thus, these generated images are not very useful for distilling the compact student model efficiently.

\begin{figure*}
  \centering
  \includegraphics[width=1\textwidth]{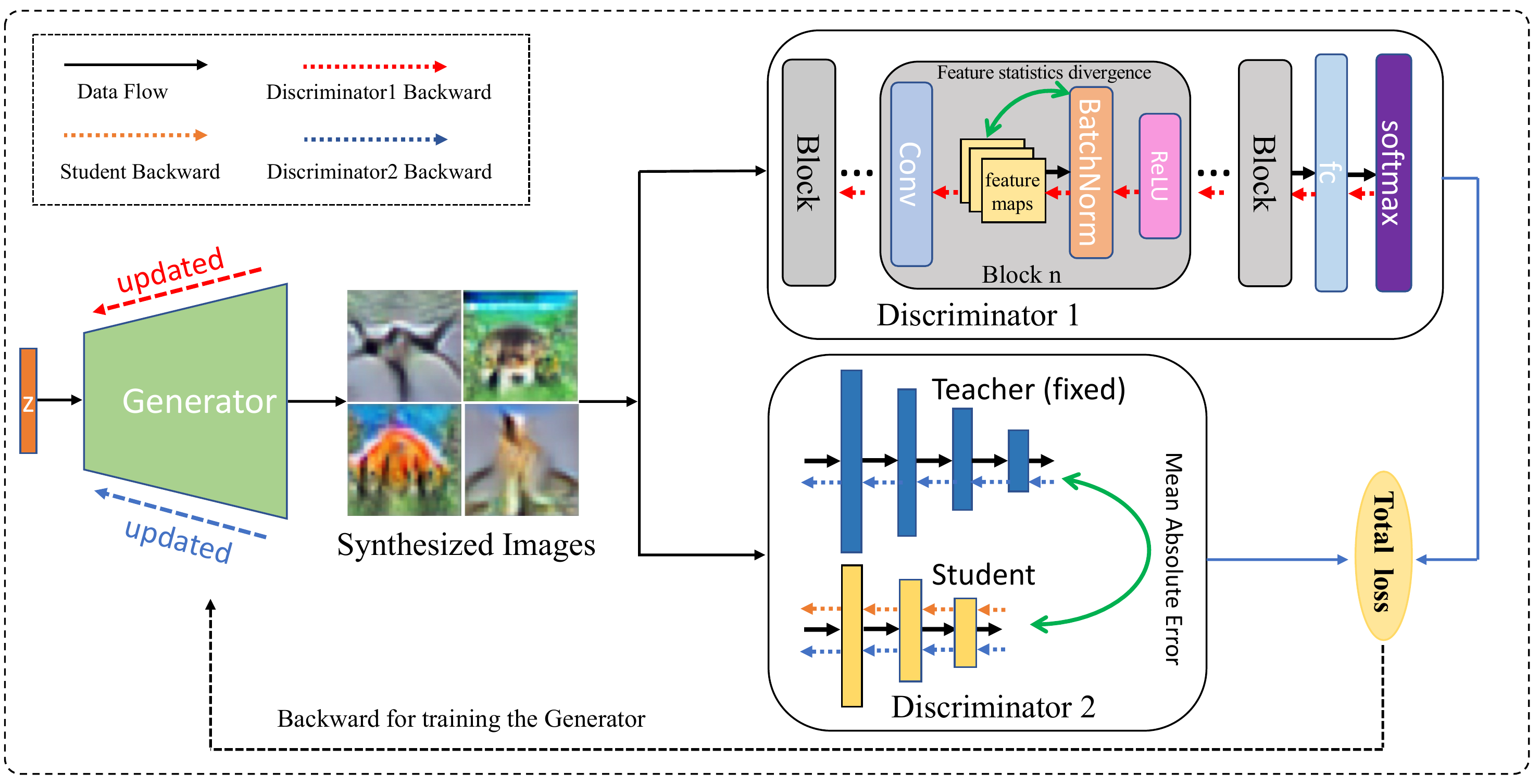}
\caption{The overall framework of the proposed dual discriminator adversarial distillation method. The training procedure is divided into two stage. First, we train a generator to produce samples instead of the original training data. Two different discriminators are employed for training the generator. Second, we use the generated images to transfer the knowledge from the teacher model to the student model.}
\label{fig:framework}
\end{figure*}

In this work, we propose a novel data-free framework to fully utilize both the intrinsic statistics from the pre-trained teacher model and the information from the student model. We train an adversarial generator through  dual  discriminator  adversarial  distillation to produce images, which mimic the training data distribution. It could also improve the generalization of the student model under the supervision of the teacher. To be specific, we use two discriminators to jointly train the generator. One fixed discriminator is the individual pre-trained teacher model which recovers the training data distribution from the existing teacher model. We utilize the valuable knowledge that encoded in the bath normalization layers which is widely used in high-performance CNNs such ResNet and DenseNet. The other is consisted of both the teacher and student model which is customized for the student to produce samples. It could improve the student model's generalization. The main contributions in this paper are summarized as follows:

\begin{itemize}
\item We introduce a novel data-free knowledge distillation framework by leveraging a two-stage dual discriminator adversarial distillation approach, which trains the portable and efficient student model without the need of training data. 

\item The proposed DDAD method not only utilizes the pre-trained teacher's intrinsic statistics in existing batch normalization layers but also obtains the maximum discrepancy from the student model.

\item Extensive experiments demonstrate the effectiveness of the proposed method on several classification and semantic segmentation benchmarks. To the best of our knowledge, this is the first work for generative based data-free knowledge distillation on large-scale datasets such as ImageNet, Cityscapes and VOC 2012.

\end{itemize}

The remaining part of this paper is organized as follows. Section 2 reviews related work and Section 3 presents our dual discriminator adversarial distillation architecture. After experimental results are presented in Section 4, we conclude this paper in Section 5.

\section{Related Works}
In this section, we briefly review some recent researches which are closely related to our approach. First, we present existing knowledge distillation methods which are divided into two categories based on the requirements of data, i.e., data-driven knowledge distillation and data-free knowledge distillation. Then, we review some mainstream methods about generative adversarial networks.

\subsection{Data-driven Knowledge Distillation}
Caruana et al. \cite{DBLP:conf/kdd/BucilaCN06} first confirm that
one ensemble of networks could transfer the knowledge to the single network. Then Ba et al. \cite{DBLP:conf/nips/BaC14} propose to teach the student network by penalizing the difference of logits between the teacher and student. Later, the concept of Knowledge Distillation (KD) is introduced by Hinton et al. \cite{Hinton2015Distilling} to solve model compression problems. It transfers the dark knowledge from the pre-trained teacher network to the compact student network by mimicking the class probabilities outputs, which are softened by setting a temperature hyperparameter in softmax. Then the performance of compact student can be retained and close to the performance of the teacher. Note that, KD requires original training data to capture the valuable knowledge from the teacher network.

However, the knowledge contained in the soft-labels is insufficient when the teacher network goes deeper. To tackle this issue, some improvements have been made, which extend KD by utilizing the intermediate representation as supervision. For example, Fitnets \cite{Romero2015FitNets} forces the student to learn the similar intermediate features as teacher's which are defined as hints. Zagoruyko et al. \cite{Zagoruyko2016Paying} define the attention maps from the intermediate features as the knowledge and then obtain a better performance compared to the one using original feature itself. Moreover, FSP \cite{Yim2017A} designs the flow distillation loss to force the student to mimic flow matrices of teacher among the feature maps between two layers. RKD \cite{park2019relational} transfers mutual relations of data examples by the distance-wise and angle-wise distillation losses. SP \cite{tung2019similarity} preserves the pairwise similarities in student's representation space instead to mimic the representation space of the teacher. CTKD \cite{9151346} combines the knowledge from different teacher models to improve the student's performance in KD. Due to excellent performance, knowledge distillation has been used to solve a variety of complex applications such as object detection \cite{DBLP:conf/iccv/DengPYZLM19,9357413}, semantic segmentation \cite{DBLP:conf/cvpr/LiuCLQLW19}, lane detection \cite{DBLP:conf/iccv/HouMLL19}, face recognition \cite{9248028,9098036,8554301} and action recognition \cite{8993763}. 

Nevertheless, the above traditional data-driven knowledge distillation methods need full of original training data, which are difficult to be obtained in real world. Thus, several few-shot knowledge distillation approaches are proposed to distill knowledge from teacher to student using few samples. Li et al. \cite{DBLP:conf/cvpr/LiL0Z20} propose to compress the teacher network by combining block-wise distillation with network pruning using few training data. Bai et al. \cite{DBLP:conf/aaai/BaiWKL20} introduce a few-shot cross distillation method which employs network pruning to reduce the layer-wise accumulated errors. The work of \cite{DBLP:conf/cvpr/WangLWG20} introduces an approach which combines mixup and active learning to distill knowledge for the student network using a small number of examples as possible. Although achieving promising results, these methods also need some original training samples and still cannot be used when the original training dataset is unavailable. Therefore, the data-free methods need to be studied for knowledge distillation.

\subsection{Data-free Knowledge Distillation}
Although most of the existing works highly rely on original training data, a few attempts have been made to distill the compact student when original training data is not accessible  \cite{DBLP:conf/cvpr/HaroushHHS20}. One kind of methods employs the generative adversarial networks to synthesize samples, which are used to distill knowledge from the teacher network to the student network. For example, Chen et al. \cite{chen2019data} propose to employ a generator to approximate the data distribution of training dataset for supervising the training of student without original training dataset. Micaelli et al. \cite{micaelli2019zero} also utilize a simple adversarial fashion to train a generator for searching images on which the student network poorly matches the teacher network. Then the generated samples are used for training the student. Yoo et al. \cite{DBLP:conf/nips/YooCKK19} propose to generate the artificial data points as original training data by training the generator and decoder networks. 

However, the methods mentioned above only take the information from the student into account, but ignore the valuable information of the pre-trained teacher model. That means the generated images from the generator are customized for the student instead of considering intrinsic statistics contained in the pre-teacher network. Thus, there has been another kind of methods for data-free knowledge distillation which obtains artificial data by updating random noise images. These methods optimize the random inputs by some regularizers while the pre-trained teacher is fixed during the training phase. For example, Nayak et al. \cite{DBLP:conf/icml/NayakMSRC19} synthesize the training samples named Data Impressions via modelling the data distribution in the softmax space. Yin et al. \cite{DBLP:conf/cvpr/YinMALMHJK20} optimize the random inputs to obtain synthesize class-conditional input images by inverting a pre-trained teacher network. During training, the knowledge contained in the batch normalization layers of the teacher is utilized for regularizing the distribution of intermediate feature maps. Haroush et al. \cite{DBLP:conf/cvpr/HaroushHHS20} extend the Inceptionism method \cite{mordvintsev2015inceptionism}, which is a feature visualization approach based on the pre-trained model, and generate samples according to low order statistics captured by existing batch normalization layers. Unfortunately, though the images from these approaches are realistic, these methods leave the student information constraint-free. That means the generator do not know what images the student network need in the distilling procedure. Moreover, the images synthesizing phase and knowledge distillation process are separated into two stages among these methods. 

\subsection{Adversarial Distillation}

Adversarial learning attracts more and more attention due to its effectiveness in generative adversarial networks (GANs) \cite{DBLP:conf/nips/GoodfellowPMXWOCB14}. It is used to capture the underlying distribution of real data sets by a generative model. Inspired by this, some adversarial knowledge distillation methods employ GANs to further promote the student model to perfectly mimic the teacher model.

Chen et al. \cite{chen2019data} propose to utilize an adversarial generator to synthesize fake samples as training data. Moreover, Liu et al. \cite{DBLP:journals/corr/abs-1812-02271} employ the generator to augment the training dataset. Micaelli et al. \cite{micaelli2019zero} further distill knowledge by generated samples from an adversarial generator. Different from the above methods, in which the teacher network is frozen during the training of the student network, Chung et al. \cite{DBLP:conf/icml/ChungPKK20} propose to jointly optimize the teacher and student in each iteration. Therefore, combining GAN and KD is an effective way to improve the student performance in KD and overcome the limitations of unaccessible data. Different from existing works, our proposed dual discriminator adversarial distillation (DDAD) employs two different discriminators to train a generator, which can produce more efficient and diversity samples for distilling knowledge from the teacher model to the student model.

\section{Proposed Method}
In this section, we will describe our DDAD method for data-free model compression in detail. As Figure \ref{fig:framework} shown, we train the generator network by employing two different discriminators in the first stage. Then we use the samples generated from the generator network to train the compact student under the supervision of teacher network. Note that the pre-trained teacher network is regarded as one discriminator, which utilizes the intrinsic statistics of existing batch normalization layer from the CNNs. The other discriminator is consisted of the teacher and the student.

\subsection{Teacher-Student Learning Paradigm}
We define the teacher model as $T$ with parameters $W_{t}$, and the student model as $S$ with parameters $W_{s}$. Furthermore, $z_{t}$ and $z_{s}$ are logits of teacher and student. There has been a temperature hyperparameter $\tau$ in vanilla knowledge distillation \cite{Hinton2015Distilling} for softening the output of the teacher model. So we define $P^\tau_{t} = softmax(z_{t}/\tau)$ and $P^\tau_{s} = softmax(z_{s}/\tau)$ as the softened output predictions of the teacher and student respectively. Then the student network is trained to mimic the behavior of the teacher network by optimizing the following loss function:

\begin{equation}
\mathcal{L}_{KD}(W_{s}) = \mathcal{H}(y_{true},P_{s}) + \lambda{KL}(P^\tau_{t},P^\tau_{s})
\label{eq.kd}
\end{equation}

\noindent where $\mathcal{H}$ is the cross-entropy loss between the true labels and outputs of student network, ${KL}$ is the Kullback-Leibler divergence between the softened outputs of teacher and student models. $\lambda$ is hyper parameter that adjust the balance between these two terms.

\subsection{Training of Generator Network}
Although GANs have the powerful ability to generate images. However, we cannot directly use vanilla GANs to generate training samples due to the absence of original ground truth in data-free scenario. Thus, we introduce a novel dual discriminator adversarial distillation to produce training samples instead of original training data. In vanilla GANs \cite{DBLP:conf/nips/GoodfellowPMXWOCB14}, the input noise vector $z$ is mapped to the desired data $x$ by the Generator $G$, i.e. $G:z \rightarrow x$, while the Discriminator $D$ is used for distinguishing the real samples from generating samples $G$ as fallowing:

\begin{eqnarray}
\mathcal{L}_{GAN} =  \mathbb{E}_{y \sim p_{data}(y)}[\log D(y)]  +
\mathbb{E}_{z \sim p_{z}(z)}[\log (1-D(G(z)))]
\end{eqnarray}

\noindent When discriminator $D$ cannot classify the image as being generated from generator $G$ or from $p_{data}$, the two player min-max optimization converges.

Furthermore, D2GAN \cite{DBLP:conf/nips/NguyenLVP17} employs two discriminators together with a generator to tackle the problem of mode collapse in GAN following the objective function:

\begin{eqnarray}
\mathcal{L} =  \alpha\mathbb{E}_{y \sim p_{data}(y)}[\log D_{1}(y)] +
\mathbb{E}_{z \sim p_{z}(z)}[ -D_{1}(G(z))] \nonumber \\
+ \mathbb{E}_{y \sim p_{data}(y)}[-D_{2}(z)] +
\beta\mathbb{E}_{z \sim p_{z}(z)}[\log (D_{2}(G(z)))]
\label{con:D2GAN}
\end{eqnarray}

\noindent By Eq. \ref{con:D2GAN}, the $D_{1}$, $D_{2}$ and G are trained alternatively. However, the discriminators of D2GAN require real training data for training. Similar to D2GAN, there is also a generator and two discriminators in our framework. The generator $\mathcal{G}$ is used to generate samples which satisfies the intrinsic statistics from the teacher and also could help the teacher to better transfer knowledge. The first discriminator $\mathcal{D}_{1}$ is the pre-trained teacher model itself with fixed parameters. And the other discriminator $\mathcal{D}_{2}$ is consisted of the teacher model and student model together.

For training the generator, we continuously optimize $\mathcal{G}$ depending on $\mathcal{D}_{1}$ and $\mathcal{D}_{2}$ by the following function:

\begin{eqnarray}
\mathcal{G} =  \mathop{\arg\min}_{G} \mathbb{E}_{z \sim p_{z}(z)}[\log (1-\mathcal{D}_{1}(G(z)))] \nonumber \\
+ \mathop{\arg\min}_{G} \mathbb{E}_{z \sim p_{z}(z)}[\log (1-\mathcal{D}_{2}(G(z)))]
\end{eqnarray}

The discriminator $\mathcal{D}_{1}$ encourages the generator to produce samples which could mimic the distribution of the original training data based on internal statistics' divergence. As we all known, almost all top-performance CNNs employ the Batch-Normalization (BN) layers which are beneficial to the performance and speed of convergence. When we feed the samples produced by the generator to $\mathcal{D}_{1}$, all BN layers calculate the running means and running variances of its inputs. Thus, we employ $\mathcal{D}_{1}$ to force the BN layers from the teacher model to mimic original means and variances stored in pre-trained teacher.

More formally, we refer the running means and variances as 
$\hat{\mu}$ and $\hat{\sigma}$, respectively. And the history means and variances stored in pre-trained teacher are defined as $\widetilde{\mu}$ and $\widetilde{\sigma}$. Then we assume that the internal statistics follow the Gaussian distribution. 

\begin{equation}
\mathcal{L}_{\mathcal{D}_{1}} = \mathbb{E}_{z \sim p_{z}(z)} {{KL}}(\mathcal{N}(\widetilde{\mu},\widetilde{\sigma}^2) || \mathcal{N}(\hat{\mu},\hat{\sigma}^2)     )\label{con:inventoryflow}
\end{equation}

In addition, we utilize the pre-trained teacher model, together with the student model to act as the discriminator $\mathcal{D}_{2}$.

\begin{equation}
\mathcal{L}_{\mathcal{D}_{2}} = - \mathbb{E}_{z \sim p_{z}(z)}[\frac{1}{n} || P^\tau_{t}(G(z))-P^\tau_{s}(G(z)) ||_{1}]
\label{con:d2}
\end{equation}

\noindent Note that, we use the negative MAE loss for this objective function. By Eq. \ref{con:d2}, we force $G$ to produce hard samples which produce the maximum discrepancy between teacher and student networks. It means that the student model do not master these samples which need to be learned. In other words, $\mathcal{D}_{2}$ is employed to customize samples for student network and get better distillation performance.

By combining the aforementioned two loss functions, i.e., Eq. \ref{con:inventoryflow} and \ref{con:d2}, we obtain the final objective function for the generator:

\begin{equation}
\mathcal{L}_{\mathcal{G}} = \delta\mathcal{L}_{\mathcal{D}_{1}} + \gamma\mathcal{L}_{\mathcal{D}_{2}}
\label{con:d}
\end{equation}

\noindent where $\delta$, $\gamma$ are hyperparameters that adjust the balance between the loss terms. By minimizing the loss function in Eq. \ref{con:d}, we train and optimize the generator. We aim to use the generator to generate samples which could mimic the data distribution of the original training data when the training loss converge.

\begin{algorithm}[t]
\caption{Dual Discriminator Adversarial Distillation} 
\label{con:algorithm}
\hspace*{-0.08cm} {\bf Input:} 
A pre-trained teacher model $T$ with parameters $W_{t}$.\\
\hspace*{-0.08cm} {\bf Output:} 
A compact student model $S$ with parameters $W_{s}$.

\begin{algorithmic}[1]
\State {Initialization: The generator $G$, the student network $S$ and training hyper-parameters.}
\State \bfseries Repeat: \mdseries
    \State \quad \bfseries Stage 1: Train the Generator. \mdseries
    \State \quad Generate vector $z$ as one batch randomly.
    \State \quad Synthesize samples as training images by $x \gets G(z)$.
    \State \quad Compute the loss function $\mathcal{L}_{\mathcal{G}}$ in Eq. \ref{con:d}.
    \State \quad Updating the weights in $G$ by back-propagation.
    
    \State \quad \bfseries Stage 2: Train the student model. \mdseries
    \State \quad Generate vector $z$ as one batch randomly.
    \State \quad Generate samples $x$ from $z$ with $G(z)$.
    \State \quad Compute the softened output of teacher network and 
    \State \quad student network: $P^\tau_{t}(x)$, $P^\tau_{s}(x)$ 
    \State \quad Compute the loss function for KD in Eq. \ref{eq.kd}.
    \State \quad Updating the weights by the gradient in $W_{s}$.
    
\State \bfseries Until: \mdseries converge.
\end{algorithmic}
\end{algorithm}

\subsection{Training of the Student Network}
In this stage, we utilize the samples produced from the generator $G$ to train the student under the supervision of the teacher model. Due to the absence of real data, the student could not get the useful knowledge from the teacher model by the soft targets. Thus, we use the MAE loss instead of KL in Eq. \ref{eq.kd}.
And the loss function for training the student is as follows:

\begin{equation}
\mathcal{L}_{\mathcal{S}} = \mathbb{E}_{z \sim p_{z}(z)}[\frac{1}{n} || P^\tau_{t}(G(z))-P^\tau_{s}(G(z)) ||_{1}]
\label{con:discriminator}
\end{equation}

Note that, we only update the student $S$ and fix the generator $G$ on this phase. The generated samples are from the generator $G$ with a batch of random noise $z$ from Gaussian distribution. 

This phase is similar to vanilla KD which distills the valuable knowledge from teacher network to student network without hesitation due to the access to real data. However, the original training data is not available in our data-free scenario. Due to the absence of ground truth labels, we cannot use the first term in Eq. \ref{eq.kd}. Thus, the generated samples by the generator are vital for our data-free distillation performance. As aforementioned, we train the generator in dual discriminator adversarial manner to produce samples which could be well used for distilling knowledge.

\subsection{Optimization}
The complete data-free distillation algorithm for our  two-stage adversarial training is summarized in Algorithm \ref{con:algorithm}. Similar to GANs, the whole training procedure can be divided into two stages.

In the first stage, we fix the discriminators and only update the generator. To be specific, we employ two different discriminators to jointly train the generator $G$, which could generate samples mimicking the original training data distribution.

Note that Eq. \ref{con:d} is the total loss function for optimizing the generator $G$. The first term of 
$\mathcal{L}_{\mathcal{G}}$ is used for discriminator $D_{1}$, which forces $G$ to generate samples satisfying the intrinsic statistics of teacher model. The second term in Eq. \ref{con:d} is for discriminator $D_{2}$ consisted with the teacher model and student model. It take into account what kind of generated samples are most useful for distilling the student model. 

In the second stage, we fix the generator $G$ and only update the student $S$ in discriminators. Note that, the student $S$, together with the teacher $T$ jointly consist of our discriminator $D_{2}$. Thus, the weights of discriminator $D_{2}$ will be updated in this stage. In addition, the pre-trained teacher network has the ability to extract semantic features, since it has been trained on the original dataset. Thus, we employ the pre-trained teacher as the fixed discriminator $D_{1}$ instead of training a new discriminator. We froze its weights in this stage.

Therefore, we optimize the student model with parameters $W_{s}$ by the KD loss in Eq. \ref{eq.kd}. Then we distill the compact student model using generated samples instead of the original training data. Our proposed method maintains diversity of generated images due to the dual discriminator adversarial distillation manner.

\section{Experiments and Results}
\label{sec:guidelines}

In the following sections, we conduct extensive experiments to verify the effectiveness of DDAD method. We implement our approach with Pytorch on NVIDIA 2080Ti GPUs. For baselines, we adopt the vanilla KD \cite{Hinton2015Distilling} method and several very recent data-free knowledge distillation works that achieve strong performance, including two non-generative based methods ZSKD \cite{DBLP:conf/icml/NayakMSRC19} and BNS \cite{DBLP:conf/cvpr/HaroushHHS20}, and some generative based data-free methods, i.e., ZSKT \cite{micaelli2019zero}, KEGNET \cite{DBLP:conf/nips/YooCKK19}, DAFL \cite{chen2019data}, DFAD \cite{DBLP:journals/corr/abs-1912-11006}, DFQ \cite{DBLP:journals/corr/abs-2005-04136} and CMI \cite{DBLP:journals/corr/abs-2105-08584}. For example, DAFL proposes to generate images that highly activate the neurons of pre-trained teacher network, and DFAD obtains a compact student network by optimizing the upper bound of the discrepancy between the student and the teacher.

\subsection{Datasets.}
\noindent\textbf{CIFAR.} The CIFAR-10 and CIFAR-100 datasets contain 60K RGB images from 10 and 100 categories, where each class has 6K and 600 images respectively. The training set contains 50K images and rest of the 10K images are used for testing set. Due to complex and advanced of ResNet \cite{DBLP:conf/cvpr/HeZRS16}, we employ ResNet-34, ResNet-18 as the teacher network and student network respectively as suggested in \cite{chen2019data}.

\noindent\textbf{Caltech101.} The Caltech101 dataset contains images of objects belonging to 101 classes, which are collected for classification. Each class contains about 50 images. We randomly divide the dataset into training set and testing set, which has 6982 images and 1695 images respectively. Note that we randomly resize and crop the images to $128\times128$ for the training procedure.

\noindent\textbf{ImageNet.} The ImageNet dataset \cite{DBLP:journals/ijcv/RussakovskyDSKS15} consists of 1.28M training images with 1000 categories and the rest of 50K images is used as the testing set. We explore the full resolution of $224\times224$ for training and testing procedures.

\noindent\textbf{CamVid.} The CamVid dataset \cite{brostow2008segmentation} is an automotive dataset for road scene segmentation. It contains 367 training and 233 testing RGB images. We use the 11 categories in CamVid such as building, tree, sky, car, road, etc. and ignore the 12th categories contains unlabeled data. We train the teacher network using the images with $128\times128$ random crop.

\noindent\textbf{NYUv2.} The NYUv2 dataset \cite{silberman2012indoor} is consisted of indoor scenes images for indoor segmentation. We use the 1449 labeled RGB-D images with 13 categories in NYUv2 and also crop the images to $128\times128$ resolution.

\noindent\textbf{Cityscapes.} The Cityscapes dataset \cite{DBLP:conf/cvpr/CordtsORREBFRS16} collects 5000 images in street scenes from 50 different cities, where 19 semantic categories are used for evaluation. It contains high quality pixel-level ﬁnely annotated images with $1024\times2048$ pixels. These images are divided into training/validation/testing with 2,975/500/1,525 images respectively. Training images for teacher are randomly cropped into $512\times512$.

\noindent\textbf{Pascal VOC.} The Pascal VOC dataset \cite{DBLP:journals/ijcv/EveringhamGWWZ10} contains 20 foreground objects categories and an extra background category. These images are divided into 1,464/1,449/1,456 groups for training/validation/testing respectively. We train the teacher network with a crop size of $512\times512$.

\subsection{Experimental Settings}
\noindent\textbf{Network architecture.}
For classification experiments, we adopt the ResNet-34 for the teacher model, the ResNet-18 for the student model, which stacks the basic residual blocks to achieve the state-of-the-art performance. It is complex and advanced to be adopted for investigating the effectiveness of our approach. For semantic segmentation experiments, the DeepLabV3 model is adopted as teacher with an ImageNet pre-tained ResNet-50 as backbone. Furthermore, we adopt the Mobilenet-V2 as backbone and train the students from scratch.

Moreover, we adopt two different generator networks for classification and semantic segmentation tasks. Specifically, we employ the generator, which uses nearest neighbor interpolation for up sampling on CIFAR dataset. For other datasets, we use a strong generator proposed in \cite{DBLP:journals/corr/RadfordMC15}, which uses the deconvolutions instead of the interpolations. In our experiments, we set the learning rate as 1e-3 and optimize the generators with Adam.

\noindent\textbf{Implementation details.}
Firstly, we train the teacher networks via the standard back-propagation. We adopt ResNet-34 as the teacher network and train it for 200 epochs using SGD with momentum 0.9. We set the weight decay to 5e-4. We use $\delta = 0.01$, $\gamma = 0.1$ and set the initial learning rate to 0.1 and then divide it by 10 at 80 and 120 epochs for CIFAR datasets. As for Caltech101, we use the same training hyperparameters and set the learning rate to 0.05 and divided by every 100 epochs. Secondly, we train student and generator networks using the proposed method. Specifically, we update the student network using SGD with momentum 0.9 and weight decay 5e-4.
For generator networks, we use Adam with a learning rate of 1e-3. The student and generator networks are trained for 500 epochs on CIFAR dataset, and 300 epochs for other datasets. For ImageNet dataset, we set the hyperparameters $\delta = 0.01$, $\gamma = 0.5$. For semantic segmentation tasks, we use the same training hyperparameters and modify the learning rate to 0.05 and reduce the weight decay to 5e-5.

Note that, we use the pre-trained ResNet-34 as the teacher network for ImageNet. Due to the mode collapse problem in such large-scale dataset, we employ multiple generators to alleviate this issue. To be specific, we employ 10 generators, each of which is trained to produce 100 classes. In another word, all the classes are divided into 100 groups, i.e., classes 1–100 in the ﬁrst group, classed 101–200 in the second group, etc. We sample 1000 images per group in a data-free manner, which uses Equation \ref{con:inventoryflow} as the optimization objective \cite{DBLP:conf/cvpr/HaroushHHS20} to get several batches of images, to estimate the per-group statistics for generator training.

\noindent\textbf{Evaluation Metrics.} For classification experiments, we take the accuracy of prediction for the metric and present the model size by the number of network parameters. We use different seeds for 3 times and adopt the median of classification accuracy as the final results. For segmentation tasks, the Intersection over Union (IoU) means the ratio of interval and union between the ground truth and the predicted segmentation for each category. In our experiments, we take the mean Intersection over Union (mIoU) for the segmentation metric. 
Furthermore, we calculate the sum of floating point operations (FLOPs) as the model complexity for one forward on a fixed input size.

\subsection{Experimental Results}
\subsubsection{CIFAR and Caltech101}
We first conduct experiments on CIFAR dataset to demonstrate the effectiveness of DDAD approach. For CIFAR dataset, we set the batch size to 256 and train the generator and student models from scratch for 500 epochs, which are randomly initialized.

The performance of classification on CIFAR-10, CIFAR-100 datasets are concluded in Table \ref{table:1}. We get the pre-trained teacher network that achieves the 95.54\%, 77.50\% accuracy on CIFAR-10 and CIFAR-100 respectively using standard back-propagation. However, the student network only achieves a 95.12\%, 76.53\% accuracy on CIFAR-10 and CIFAR-100 when trained individually, which are obviously lower than those of the pre-trained teacher network. Then we train the student network using the vanilla knowledge distillation, which improves the classification accuracy of the student model from 95.12\%, 76.53\% to 95.43\% 76.87\% on CIFAR-10 and CIFAR-100 respectively using the original training data.

The rest rows of Table \ref{table:1} reports the classification results of several recent data-free knowledge distillation approaches. These methods could be divided into two kinds, i.e., non-generative based data-free methods and generative based data-free methods. We compare our approach with the state-of-the-art data-free methods using their released codes. First, we compare our method with two non-generative based methods ZSKD \cite{DBLP:conf/icml/NayakMSRC19} and BNS \cite{DBLP:conf/cvpr/HaroushHHS20}. As can be seen from Table \ref{table:1}, ZSKD and BNS achieve test accuracies of 91.61\% and 93.02\% on CIFAR-10, 70.21\% and 73.25\% on CIFAR-100 respectively. The other generative based data-free methods, i.e., ZSKT \cite{micaelli2019zero}, KEGNET \cite{DBLP:conf/nips/YooCKK19}, DAFL \cite{chen2019data}, DFAD \cite{DBLP:journals/corr/abs-1912-11006}, DFQ \cite{DBLP:journals/corr/abs-2005-04136} and CMI \cite{DBLP:journals/corr/abs-2105-08584}, achieve the test accuracies of 85.95\%, 91.83\%, 92.22\%, 93.30\%, 93.26\% and 94.08\% on CIFAR-10 dataset, 66.29\%, 73.91\%, 74.47\%, 69.43\%, 67.01\% and 74.01\% on CIFAR-100 dataset. The results of our proposed DDAD method are shown in the bottom of Table \ref{table:1}. We obtain the classification accuracy of 94.81\% on CIFAR-10 dataset, which is 2.59\% and 1.51\% better than DAFL's and DFAD's performance. This is very close to 95.43\% which is obtained in KD using the original training data. For CIFAR-100, the accuracy of our method is 75.04\%, exceeding the DAFL and DFAD by 0.57\% and 5.61\%.

\begin{table*}\tiny 
\begin{center}

\caption{Classification accuracy (\%) on CIFAR-10, CIFAR-100, Caltech101 and ImageNet datasets (3 runs). We use the standard back-propagation to train the student individually. DDAD means the performance of the student model from our method.}
\label{table:1}
\begin{tabular}{c|c|c|c|c|c|c}
\hline
Algorithm & Network & Required data & CIFAR-10 &  CIFAR-100  & Caltech101 & ImageNet \\
\hline
\hline

Student
& ResNet-18 & \checkmark & 95.12\% & 76.53\% & 75.69\% & 69.64\%\\

\cline{2-7}
Teacher
& ResNet-34 & \checkmark & 95.54\% & 77.50\% & 76.60\% & 73.27\%\\

\cline{2-7}
Knowledge Distillation
& ResNet-18 & \checkmark & 95.43\% & 76.87\% & 76.12\% & 70.87\%\\
\hline
\hline

Zero-Shot KD (ZSKD) \cite{DBLP:conf/icml/NayakMSRC19} 
& ResNet-18 & -- & 91.61\% & 70.21\% & 69.01\% & --\\

\cline{2-7}
Knowledge within (BNS) \cite{DBLP:conf/cvpr/HaroushHHS20}
& ResNet-18 & -- & 93.02\% & 73.25\% & 72.71\% & 51.32\%\\
\hline
\hline

Zero-Shot Transfer(ZSKT)\cite{micaelli2019zero}
& ResNet-18 & -- & 85.95\% & 66.29\% & 65.82\% & --\\

\cline{2-7}
KEGNET \cite{DBLP:conf/nips/YooCKK19}
& ResNet-18 & -- & 91.83\% & 73.91\% & 72.29\% & --\\

\cline{2-7}
Data-Free Learning (DAFL) \cite{chen2019data}
& ResNet-18 & -- & 92.22\% & 74.47\% & -- & --\\

\cline{2-7}
Data-free Adversarial (DFAD) \cite{DBLP:journals/corr/abs-1912-11006}
& ResNet-18 & -- & 93.30\% & 69.43\% & 73.50\% & 20.63\%\\

\cline{2-7}
Data-Free Quantization (DFQ) \cite{DBLP:journals/corr/abs-2005-04136}
& ResNet-18 & -- & 93.26\% & 67.01\% & 73.61\% & --\\

\cline{2-7}
Contrastive Inversion (CMI) \cite{DBLP:journals/corr/abs-2105-08584}
& ResNet-18 & -- & 94.08\% & 74.01\% & 73.92\% & --\\

\cline{2-7}
Our DDAD
& ResNet-18 & -- & \bfseries94.81\%\mdseries & \bfseries 75.04\% \mdseries  & \bfseries75.01\%\mdseries & \bfseries59.84\%\mdseries\\
\hline
\end{tabular}
\end{center}
\end{table*}

Furthermore, we conduct experiments on the complex Caltech101 dataset which has 101 classes and images of size $128\times128$. The teacher network obtains the accuracy of 76.60\% and the student network achieves the accuracy of 75.69\% when trained individually. DFAD achieves only 73.50\% accuracy without using any original training data and DAFL fails in this case. Our approach gets the highest accuracy among these data-free methods. The student network trained using our method achieves 75.01\% accuracy, which is comparable with that of the pre-trained teacher network.

\subsubsection{ImageNet}
To further investigate the effectiveness of our method, we extend our method to ImageNet dataset. We set the batch size to 128 and train the ensemble of 10 generators and student network from scratch for 100K steps.

We present the testing accuracy of classification on ImageNet in Table \ref{table:1}. The pre-trained weights from torchvision are used for the teacher network which achieves an accuracy of 73.27\%. However, the student network only achieves a 69.64\% accuracy when trained individually. We first use the vanilla knowledge distillation based on data-driven to train the student, which outperforms the student trained individually by 1.23\%. Then we compare our method with the state-of-the-art data-free approaches. To the best of our knowledge, there are few previous methods that work for data-free distillation on ImageNet, especially for generative based methods. We compare our method with BNS \cite{DBLP:conf/cvpr/HaroushHHS20} and DFAD \cite{DBLP:journals/corr/abs-1912-11006}, which belong to non-generative and generative based data-free distillation methods respectively. Note that, the other state-of-the-art baselines fail on such large-scale dataset.

\begin{figure}
  \centering
  \includegraphics[width=8.5cm]{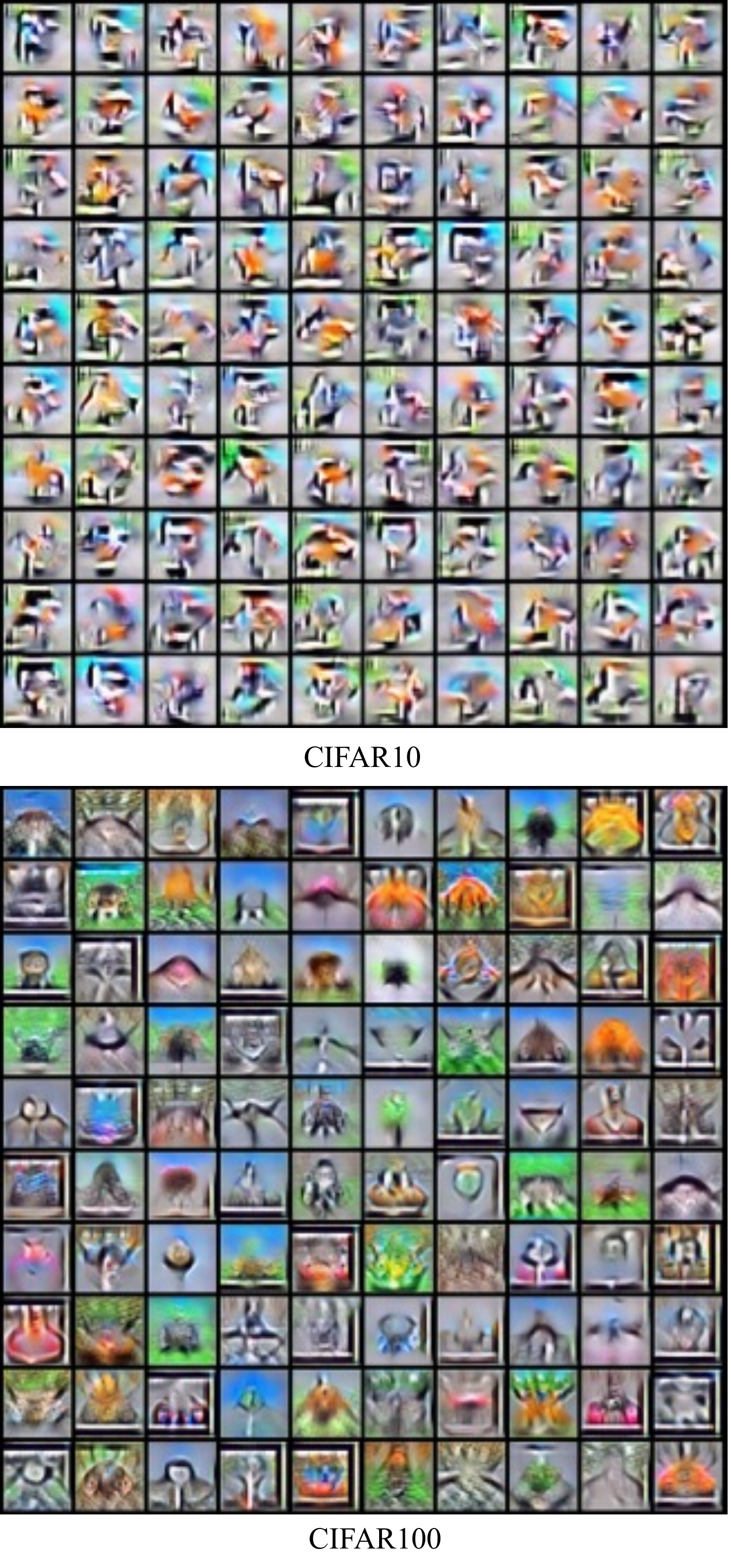}
\caption{Generated samples on CIFAR-10 and CIFAR-100. Given only a pre-trained ResNet-34 teacher network, a randomly initialized ResNet-18 student network.}
\label{fig:2}
\end{figure}

As can be seen from Table \ref{table:1}, the student of the proposed DDAD method achieves an accuracy of 59.84\%, which outperforms the ones trained by BNS and DFAD by 8.52\% and 39.21\% respectively under the same steps and setting.

\par\textbf{Visualization of Generated Samples.} We present a set of generated samples by applying our method to a CIFAR-10/CIFAR-100 pre-trained ResNet-34 in Figure \ref{fig:2}. Note that, these images are generated from the generators under our framework. Comparing with the generated samples on CIFAR-10 and CIFAR-100, we can ﬁnd that the generator on CIFAR-100 produces more complicated samples than that on CIFAR-10. As the difﬁculty of classiﬁcation increases, the teacher model becomes more knowledgeable so that, in adversarial learning, the generator can recover more complicated images. Furthermore, our method produces some images features and textures, e.g., the flowers, airplanes and cars as presented in the right of Figure \ref{fig:2}. Meanwhile, we could observe that the generated images are almost different from each other. We owe the perfect image diversity to the dual discriminators used in our framework.

\begin{figure}
  \centering
  \includegraphics[width=1\textwidth]{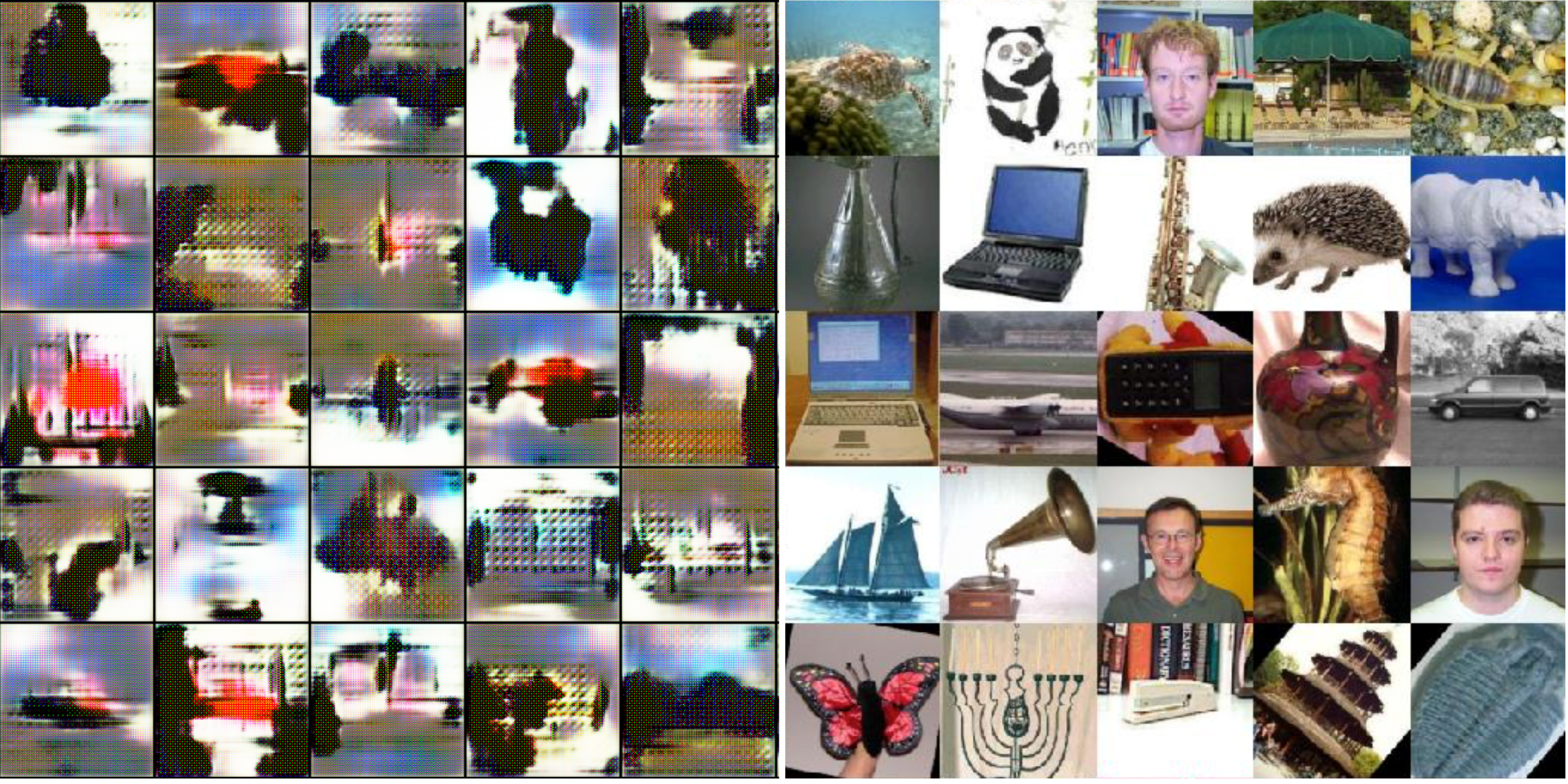}
\caption{Generated samples (left) of size $128\times128$ and real samples (right) from Caltech101. Given only a Caltech-101 pre-trained ResNet-34 teacher network and a randomly initialized ResNet-18 student network.}
\label{con:caltech101}
\end{figure}

\begin{figure}
  \centering
  \includegraphics[width=1\textwidth]{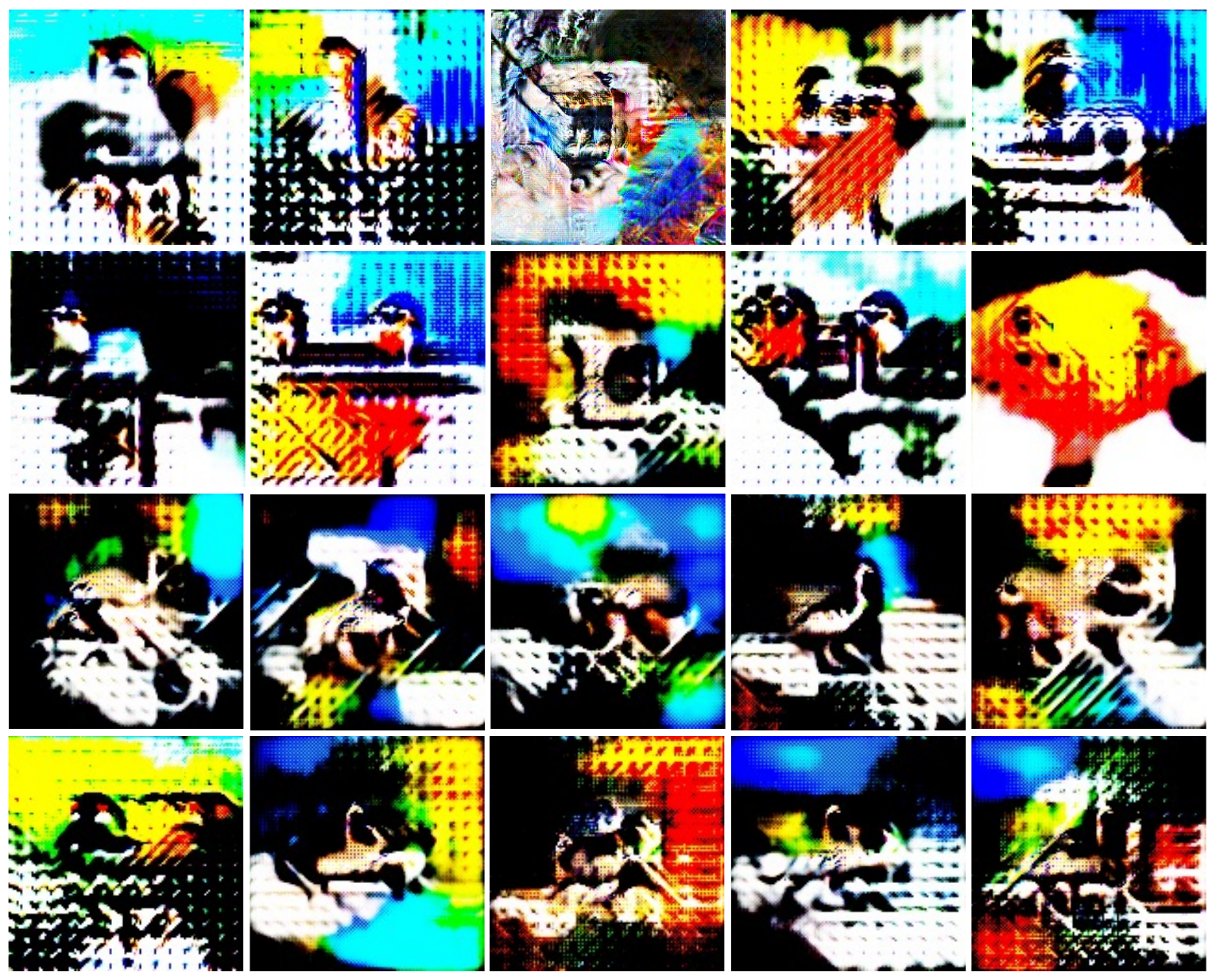}
\caption{Generated samples of size $224\times224$ on ImageNet. Given only an ImageNet pre-trained ResNet-34 teacher network and a randomly initialized ResNet-18 student network.}
\label{con:ImageNet}
\end{figure}

The left of Figure \ref{con:caltech101} shows the generated samples of size $128\times128$ by applying the proposed approach to a Caltech-101 pre-trained ResNet-34. And the right side of Figure \ref{con:caltech101} shows the real samples from the original training dataset, which are used to train the ResNet-34 teacher network. As can be seen from Figure \ref{con:caltech101}, the generated images, which could not be recognized by humans, significantly improve the performance of student model. Hence, it implies that the realistic images are vital but not the indispensable way for knowledge distillation.

We also present the samples produced from the generators using the proposed DDAD method in Figure \ref{con:ImageNet}. Figure \ref{con:ImageNet} shows that the images produced by the generators of our method seem to be less closer to the original training images, but indeed have abundant knowledge for the distillation task. In other words, due to lack of original training images, the generators of our DDAD method are unable to produce the images which can fool humans' eyes as real ones, but they could produce the real samples by the deep neural networks' perception. It also indicates the gap between deep neural networks and the human.

\subsubsection{CamVid and NYUv2}

\begin{table*}[t]\small
\begin{center}
\caption{The mIoU of DeepLabv3 model on CamVid, NYUv2, Cityscapes and VOC 2012 datasets. The teaher model are pre-trained on ImageNet when the student are randomly intialized. }
\label{table:2}
\begin{tabular}{c|c|c|c|c|c|c|c|c}
\hline
 & \multicolumn{2}{|c|}{CamVid} & \multicolumn{2}{|c}{NYUv2} & \multicolumn{2}{|c}{Cityscapes} & \multicolumn{2}{|c}{VOC 2012}  \\
\hline
\hline
Method & FLOPs & mIoU &  FLOPs & mIoU & FLOPs & mIoU &  FLOPs & mIoU\\
\hline
\hline
Teacher &  41.0G & 0.819  & 41.0G & 0.731 & 51.4G & 0.749 & 51.4G & 0.720\\

ZSKT \cite{micaelli2019zero} & 5.54G& 0.210 &5.54G & 0.173 & 6.0G & 0.063 & 6.0G & 0.108\\

DAFL \cite{chen2019data} &  5.54G & 0.010 & 5.54G & 0.105  &6.0G & 0.301 &6.0G& 0.293\\

DFAD \cite{DBLP:journals/corr/abs-1912-11006} &  5.54G & 0.494 & 5.54G & 0.319  &6.0G & 0.394 &6.0G& 0.357\\
\hline
\hline
Ours &  5.54G & \bfseries 0.518 \mdseries & 5.54G & \bfseries 0.363 \mdseries  &6.0G & \bfseries0.424\mdseries &6.0G & \bfseries0.407\mdseries \\
\hline
\end{tabular}
\end{center}
\end{table*}

\begin{figure*}
  \centering
  \includegraphics[width=1\textwidth]{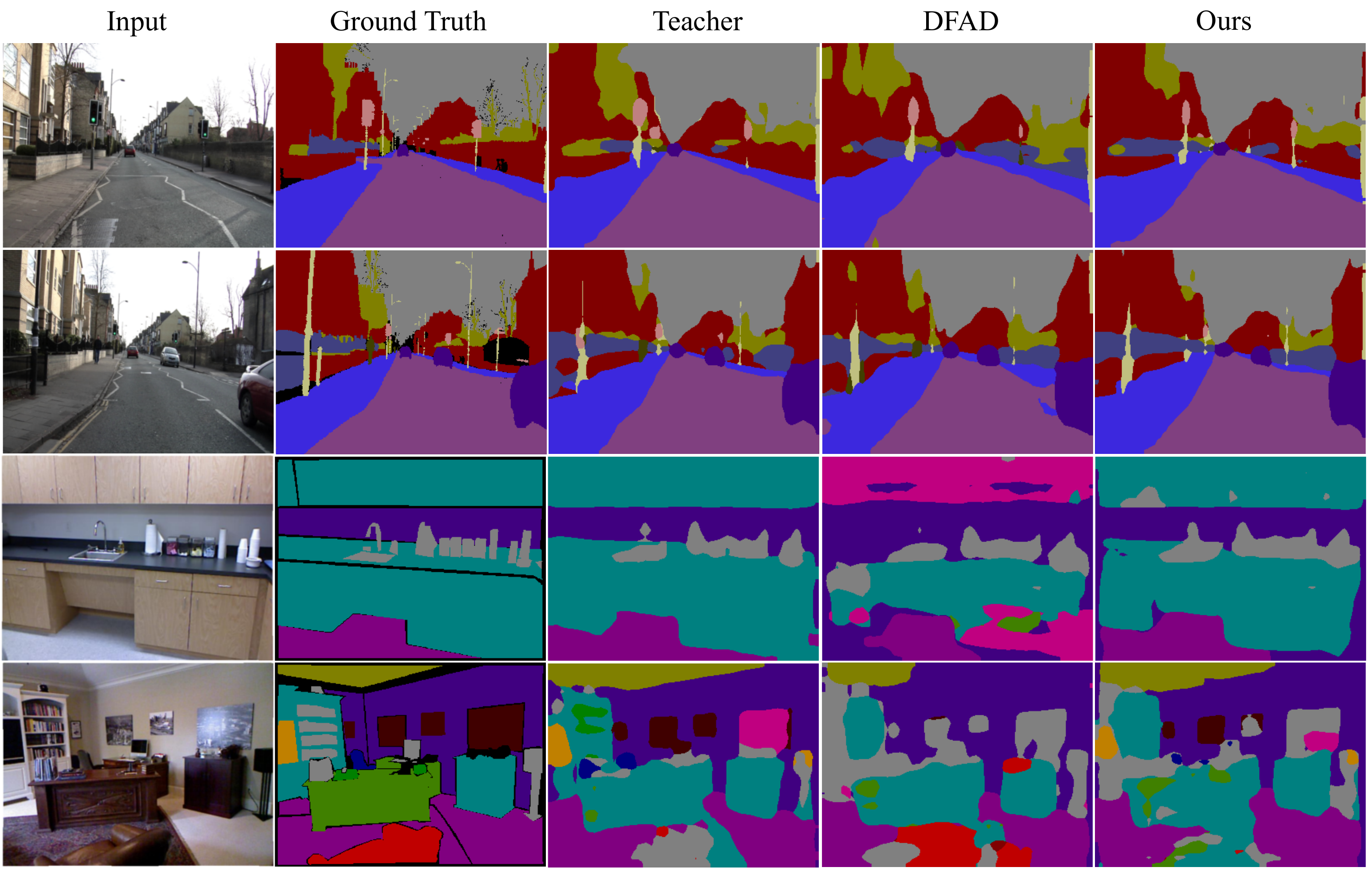}
\caption{Qualitative segmentation results on CamVid and NYUv2 produced from MobileNet-V2. Given only the ImageNet pre-trained ResNet-50, which is adopted as backbone. The segmentation results are obtained on generated samples. Our method achieves the best performance when the original training data is not available. Moreover, our method is comparable with that of the teacher network.}
\label{con:seg}
\end{figure*}

In addition to verifying our method on the above described classification task, we also extend the proposed approach to semantic segmentation tasks. We use the resolution of $128 \times 128$ cropped images for CamVid and NYUv2 in the following experiments. The teacher model is initialized by the pre-trained ResNet-50 on ImageNet dataset. Furthermore, we train the student network from scratch using the generated samples with the resolution of $128 \times 128$.
The batch size is set to 32 in our experiments. For CamVid dataset, we use SGD with a learning rate of 0.1 and a weight decay of 5e-4. The generator and student networks are trained for 300 epochs and the learning rate is decayed every 100 epochs. For NYUv2, we set the learning rate as 0.05 and we multiply it by 0.3 on 150 epochs and 250 epochs. We also train the generator and student networks for 300 epochs and the weight decay is reduced to 5e-5.

Table \ref{table:2} presents the results of the student model (Mobilenet-V2) on our framework and other state-of-the-art results. We assess the complexity by Flops that is computed on the resolution $128 \times 128$. As can be seen from Table \ref{table:2}, our approach yields mIoU 51.8\% on CamVid dataset, which boosts the performance of DFAD by 2.4 point. For NYUv2 dataset, our method achieves 36.3\% mIoU, which improves the performance of DFAD by 4.4\%.

Furthermore, we conduct the visualization experiments to verify the effectiveness of our dual discriminator adversarial distillation method and show the qualitative segmentation results are shown in Figure \ref{con:seg}. As can be seen from Figure \ref{con:seg}, some objects such as traffic signs are clearer than those in DFAD \cite{DBLP:journals/corr/abs-1912-11006} method and even close to the performance of teacher.

\begin{figure*}
  \centering
  \includegraphics[width=1\textwidth]{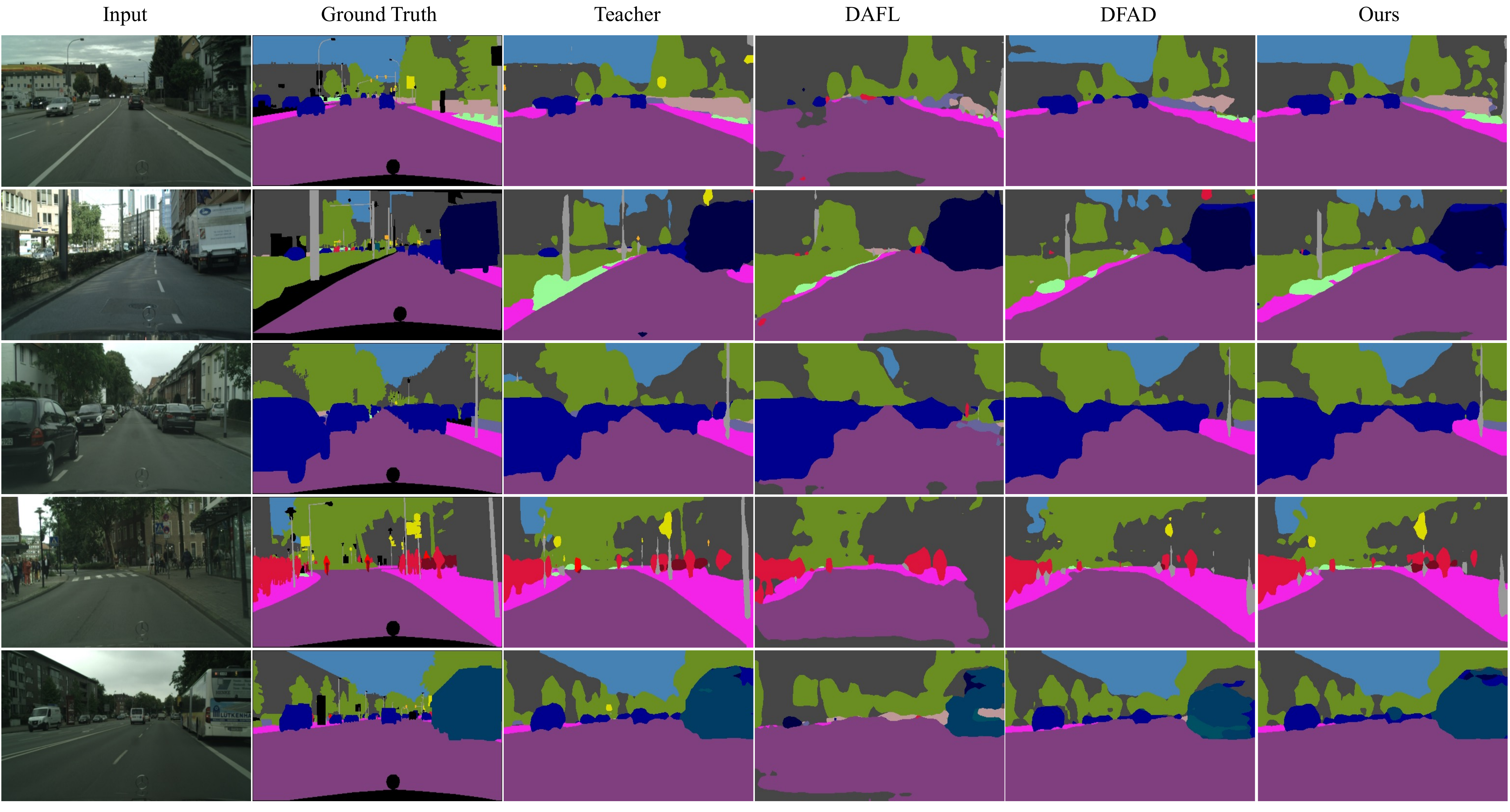}
\caption{Qualitative segmentation results on CityScapes produced from MobileNet-V2. Given only the ImageNet pre-trained ResNet-50, which is adopted as backbone. The segmentation results are obtained on generated samples. Our method achieves the best performance when the original training data is not available. Moreover, our method is comparable with that of the teacher network.}
\label{con:seg02}
\end{figure*}

\subsubsection{Cityscapes and VOC 2012}
To further verify the effectiveness of our method, we
extend our method to the large-scale and challenging semantic segmentation datasets, such as Cityscapes and VOC 2012. In the following experiments, we use the resolution of $512\times512$ cropped images for Cityscapes and VOC 2012. Furthermore, we train the student model from scratch using the images with the resolution of $512\times512$ produced by the trained generators.

We also list the results of the student network under our method and other state-of-the-art baselines in Table \ref{table:2}. The complexity is assessed by Flops that is computed on the resolution $512\times512$. As can be seen, our approach achieves 42.4\% mIoU on Cityscapes, which gains the improvements of DFAD, DAFL and ZSKT by 3\%, 12.3\% and 36.1\%, respectively. For VOC dataset, our method yields mIoU 40.7\%, which boosts the performance of DFAD, DAFL and ZSKT by 5\%, 11.4\% and 29.9\%, respectively.

\begin{figure*}
  \centering
  \includegraphics[width=1\textwidth]{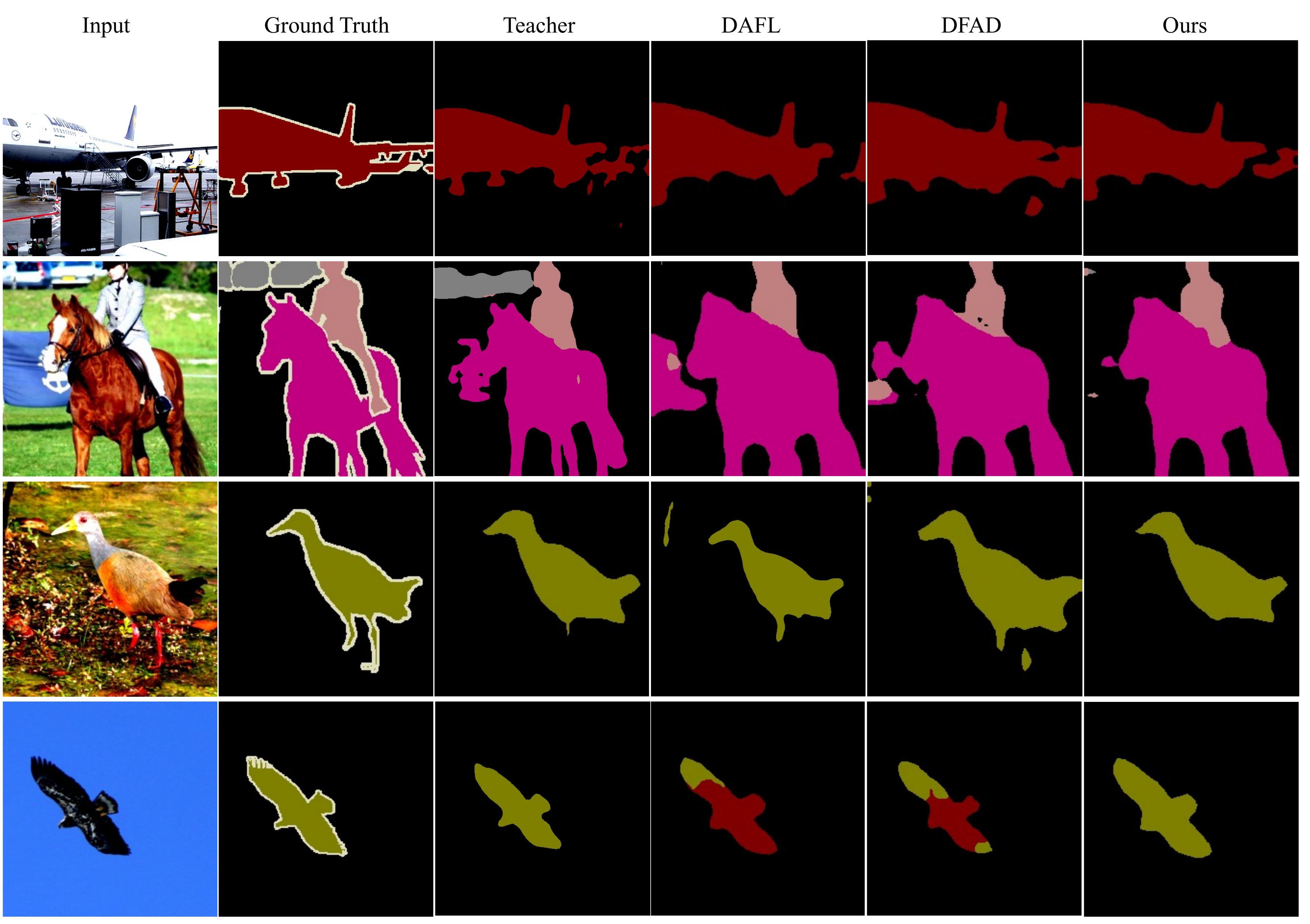}
\caption{Qualitative segmentation results on VOC 2012 produced from MobileNet-V2. Given only the ImageNet pre-trained ResNet-50, which is adopted as backbone. The segmentation results are obtained on generated samples. Our method achieves the best performance when the original training data is not available. Moreover, our method is comparable with that of the teacher network.}
\label{con:seg03}
\end{figure*}

\subsection{Ablation Study}
Below we perform several extended experiments on CIFAR dataset to verify the performance of our DDAD method. To investigate the robustness of DDAD, we consider discussing the effectiveness of each discriminator in our framework and distill student network which has the same architecture as the teacher network using our approach.

\noindent\textbf{The effectiveness of each discriminator.} We look into the effect of each discriminator in our dual discriminator adversarial distillation framework. And we conduct the following experiments on CIFAR-100. And we employ the ResNet-34 as the teacher model, ResNet-18 as the student model. As can be seen from Eq. \ref{con:d}, $\delta$ and $\gamma$ are hyperparameters that adjust the balance between two discriminators. Thus we range the hyperparameters $\delta$ and $\gamma$ from 0.005 to 0.025, 0.01 to 0.2 respectively. The defaults of $\delta$ and $\gamma$ are 0.01 and 0.1 respectively. When we change one of the hyperparameters, the other keeps the default. 

Table \ref{table:3} presents that the classification results of DDAD approach are affected by setting different hyper-parameters. As can be seen from Table \ref{table:3}, the accuracy performance is insensitive to the choice of $\gamma$ over a large range from 0.05 to 0.15, 0.4 to 0.6 on CIFAR100/Caltech101 and CamVid datasets, respectively. However, a carefully adjusted hyperparameters do bring the best performance. When we set one of the hyperparameters to 0, it means that we use the single discriminator in our framework. Experimental results show that both of the discriminators in our framework bring performance improvement.

\begin{table*}\tiny
\begin{center}
\caption{Effects of hyper-parameters $\delta$, $\gamma$ in DDAD on CIFAR-100, Caltech101 and CamVid datasets. For classification task, we use the ResNet-34 for the teacher model, the ResNet-18 for the student model. For segmentation task, we adopt DeepLabV3 model as teacher with a pre-trained ResNet-50 as backbone, and MobileNet-V2 as student.}
\label{table:3}
\begin{tabular}{c|c|c|c|c|c|c|c|c|c|c|c|c}
\hline

\multirow{2}* {\tabincell{c}{CIFAR100}}&
\multicolumn{6}{c|}{$\delta$} & \multicolumn{6}{c}{$\gamma$}\\

\cline{2-13} 
   & 0 & 0.005 & 0.01 & 0.015 & 0.02 & 0.025 & 0 & 0.01 & 0.05 & 0.1 & 0.15 & 0.2 \\ 
\hline
Acc (\%)& 67.63 & 69.52 & 75.04 & 69.92 & 68.43 & 68.07 & 67.92 & 69.87 & 74.42 & \bfseries 75.04 \mdseries & 74.90 & 68.13\\
\hline
\hline

\multirow{2}* {\tabincell{c}{Caltech101} }&
\multicolumn{6}{c|}{$\delta$ } & \multicolumn{6}{c}{$\gamma$ }\\
\cline{2-13} 
 & 0 & 0.005 & 0.01 & 0.015 & 0.02 & 0.025 & 0 & 0.01 & 0.05 & 0.1 & 0.15 & 0.2  \\
\hline
Acc (\%) & 68.01 & 72.39 & 75.01 & 73.17 & 70.68 & 68.71 & 68.63 & 70.37 & 73.91 & \bfseries 75.01 \mdseries & 73.70 & 69.92\\
\hline
\hline

\multirow{2}* {\tabincell{c}{CamVid} }&
\multicolumn{6}{c|}{$\delta$} & \multicolumn{6}{c}{$\gamma$ }\\
\cline{2-13} 
 & 0 & 0.005 & 0.01 & 0.015 & 0.02 & 0.025 & 0 & 0.3 & 0.4 & 0.5 & 0.6 & 0.7  \\
\hline
mIoU (\%) & 46.7 & 48.6 & 51.8 & 47.3 & 45.8 & 44.9 & 43.9 & 44.7 & 46.9 & \bfseries 51.8 \mdseries & 47.5 & 45.4\\
\hline

\end{tabular}
\end{center}
\end{table*}

To be specific, when we set the hyperparameter $\delta$ to $0$ as shown in Table \ref{table:3}, it means only the discriminator $D_{2}$ is employed in our framework. In such a situation, the student network only achieves the accuracy of 67.63\%, since the generator is trained only for synthesizing samples. It fools the student but ignores the intrinsic statistics of the pre-trained teacher. On the contrary, when we set $\gamma$ to $0$ in Table \ref{table:3}, it means that we only employ the discriminator $D_{1}$ in our framework. Note that, the discriminator $D_{1}$ is fixed during training in this situation. Thus, we only train and update the generator $G$ in the whole training process. And the distillation process of the student is followed in the next stage. As presented in Table \ref{table:3}, the student only achieves the accuracy of 67.92 \% in such a situation. In addition, it also can be time-consuming to distill the student network.

\begin{table}
\begin{center}
\caption{The classification on several public datasets. Given only the pre-trained ResNet-34 teacher networks. Note that the student has the same network architecture as the teacher network.}
\label{table:4}
\begin{tabular}{c|c|c|c}
\hline
 Dataset & Network & \multicolumn{2}{|c}{Accuracy}  \\
\hline
 &  & Teacher &  Student\\
\hline
\hline
CIFAR-10 &  ResNet-34 & 95.54  & 93.82 \\
CIFAR-100 &  ResNet-34 & 77.50 & 75.79 \\
Caltech101 &  ResNet-34 & 76.60 & 74.96 \\
\hline
\end{tabular}
\end{center}
\end{table}

\noindent\textbf{Difference between generated samples and real data.} To further explore the effectiveness of our method, we compare the classification results between the generated samples and original training data. Thus we adopt the same network architecture of teacher as the student network in our framework. The performance is presented in Table \ref{table:4}.

As can be seen from Table \ref{table:4}, the ResNet-34 networks obtain a 95.54\% accuracy and a 77.50\% accuracy on CIFAR-10 and CIFAR-100 datasets, respectively. However, the student networks, which also use the same ResNet-34 networks as the teacher network, achieve the accuracy of 93.82\% and 75.79\%, respectively. Thus, we conclude that the performances of students are very close to those of the teacher models. For Caltech101 dataset the student also achieves very comparable performance to those of teacher networks. These  results imply that the generated samples from our framework could well mimic the original training data and improve the performance of student network in knowledge distillation.

\section{Conclusion}
In this paper, we propose a novel data-free knowledge distillation framework to obtain a compact but efficient student model without the need of the original training data. This is achieved by designing a dual discriminator adversarial distillation approach in two stages. In the first stage, we employ two different discriminators to train the generator which could produce images that instead of the original training data. Moreover, because of the dual discriminator, the generated samples could mimic the original training data distribution. Here, we not only use the teacher model's intrinsic statistics but also get the maximum discrepancy on the student model to mimic the data distribution. In the second stage, we distill knowledge from the teacher to the student model with the generated samples from the first stage. Experimental results on several classification and segmentation benchmarks demonstrate that the proposed approach yields state-of-the-art data-free distillation performance, outperforming existing data-free knowledge distillation methods. In the future, we will explore the prior knowledge to further improve the diversity of synthetic images on the proposed dual discriminator adversarial distillation framework.

\section*{Acknowledgment}
We thank supports of the National Natural Science Foundation of China (No. 61971388, U1706218, L1824025), and Key Natural Science Foundation of Shandong Province (No. ZR2018ZB0852).

\section*{Data availability statement}
The datasets generated during and/or analysed during the current study are available in \href{https://github.com/ouc-ocean-group}{https://github.com/ouc-ocean-group.}

%\begin{acknowledgements}
%If you'd like to thank anyone, place your comments here
%and remove the percent signs.
%\end{acknowledgements}

% Authors must disclose all relationships or interests that 
% could have direct or potential influence or impart bias on 
% the work: 
%
% \section*{Conflict of interest}
%
% The authors declare that they have no conflict of interest.

% BibTeX users please use one of
\bibliographystyle{spbasic}      % basic style, author-year citations

\bibliography{mybibfile}

\end{document}